\documentclass{ieeeaccess}
\usepackage{cite}
\usepackage{amsmath,amssymb,amsfonts}
\usepackage{blindtext}
\usepackage{enumerate}
\usepackage{footnote}
\usepackage{url}
\usepackage{graphicx}
\usepackage{float}
\usepackage{textcomp}
\usepackage{comment}
\usepackage{color,soul}
\usepackage{textcomp}
\usepackage{algorithm}
\usepackage{graphicx,wrapfig}
\usepackage{algorithmic}

\usepackage[table,xcdraw,usenames,dvipsnames]{xcolor}
\usepackage{array}
\newcolumntype{P}[1]{>{\centering\arraybackslash}p{#1}}
\usepackage[colorlinks=true]{hyperref}
\usepackage{amsthm}   
\usepackage{cleveref}
\hypersetup{
    colorlinks=true,
    linkcolor=blue,
    filecolor=blue,      
    urlcolor=blue,
    citecolor=blue,
}
\usepackage{lineno}

\usepackage{tikz}
\NewSpotColorSpace{PANTONE}
  \AddSpotColor{PANTONE} {PANTONE3015C} {PANTONE\SpotSpace 3015\SpotSpace C} {1 0.3 0 0.2}
  \SetPageColorSpace{PANTONE}

\definecolor{lime}{HTML}{A6CE39}
\DeclareRobustCommand{\orcidicon}{
	\begin{tikzpicture}
	\draw[lime, fill=lime] (0,0) 
	circle [radius=0.16] 
	node[white] {{\fontfamily{qag}\selectfont \tiny ID}};
	\draw[white, fill=white] (-0.0625,0.095) 
	circle [radius=0.007];
	\end{tikzpicture}
	\hspace{-2mm}
}

\foreach \x in {A, ..., Z}{\expandafter\xdef\csname orcid\x\endcsname{\noexpand\href{https://orcid.org/\csname orcidauthor\x\endcsname}
			{\noexpand\orcidicon}}
}

\def\BibTeX{{\rm B\kern-.05em{\sc i\kern-.025em b}\kern-.08em
    T\kern-.1667em\lower.7ex\hbox{E}\kern-.125emX}}

\begin{document}
\history{Date of publication 07/09/2021}
\doi{10.1109/ACCESS.2021.3111005}

\title{HSMD: An object motion detection algorithm using a Hybrid Spiking Neural Network Architecture}

\author{\uppercase{Pedro Machado}\authorrefmark{1} \IEEEmembership{Student Member, IEEE}\orcidA{},
\uppercase{Andreas Oikonomou}\authorrefmark{1} \orcidB{}, \uppercase{Jo\~ao Filipe Ferreira}\authorrefmark{1} \IEEEmembership{Member, IEEE} \orcidC{} and \uppercase{T.M. McGinnity}\authorrefmark{2} \orcidD{}
\IEEEmembership{Senior Member, IEEE}}
\address[1]{Computational Neurosciences and Cognitive Robotics Group (ISTeC031), School of Sciences and Technology, Nottingham Trent University, Clifton Campus, Nottingham, NG11 8NS, United Kingdom (e-mails: \{pedro.baptistamachado,andreas.oikonomou,joao.ferreira\}@ntu.ac.uk}
\address[2]{Intelligent Systems Research Centre (MS225), Ulster University, Magee Campus, BT48 7JL, Derry, N. Ireland (e-mail: tm.mcginnity@ulster.ac.uk)}

\markboth
{Machado P. \headeretal: HSMD: An object motion detection algorithm...}
{Machado P. \headeretal: HSMD: An object motion detection algorithm...}

\corresp{Corresponding author: Pedro Machado (e-mail: pedro.baptistamachado@ntu.ac.uk).}

\begin{abstract}
\newline
\begin{Huge}
\end{Huge}
\newline
The detection of moving objects is a trivial task performed by vertebrate retinas, yet a complex computer vision task. Object-motion-sensitive ganglion cells (OMS-GC) are specialised cells in the retina that sense moving objects. OMS-GC take as input continuous signals and produce spike patterns as output, that are transmitted to the Visual Cortex via the optic nerve. The Hybrid Sensitive Motion Detector (HSMD) algorithm proposed in this work enhances the GSOC dynamic background subtraction (DBS) algorithm with a customised 3-layer spiking neural network (SNN) that outputs spiking responses akin to the OMS-GC. The algorithm was compared against existing background subtraction (BS) approaches, available on the OpenCV library, specifically on the 2012 change detection (CDnet2012) and the 2014 change detection (CDnet2014) benchmark datasets. The results show that the HSMD was ranked overall first among the competing approaches and has performed better than all the other algorithms on four of the categories across all the eight test metrics. Furthermore, the HSMD proposed in this paper is the first to use an SNN to enhance an existing state of the art DBS (GSOC) algorithm and the results demonstrate that the SNN provides near real-time performance in realistic applications.
\end{abstract}


\maketitle

\section{Introduction} \label{ch1:introduction}

The retina is a tiny tissue of about 1mm depth at the back of the eye, and it is responsible for the first stage of biological image processing. All vertebrate retinas possess a variable number of the same type of retinal cells, namely, photoreceptors (rods and cones), horizontal, bipolar, amacrine and ganglion cells \cite{Kolb2003,Holmes2018}. Light stimuli are sensed by the photoreceptors, which trigger electrical and chemical signals that propagate through the retinal cells and are transported via the optic nerve to the Visual Cortex. Retinal photoreceptors are sensitive to dim light (rods), colour vision (cones) and bright light (cones), and connect to bipolar and horizontal cells. Horizontal cells are responsible for regulating the signals triggered by neighbouring rods and cones. Bipolar cells receive process and transmit signals from groups of rods and cones to ganglion cells. Amacrine cells interact with groups of neighbouring bipolar cells to regulate signals transmitted to the ganglion cells, responsible for collecting the visual signals and propagating them to the visual cortex via millions of parallel channels in the optic nerve \cite{Kolb2003,Holmes2018}. The types of retinal cells vary in concentration, functionality and size. There are thousands of retinal circuits formed by types and sub-types of retinal cells wired together \cite{Kolb2003}. Different retinal circuits trigger different functionalities such as light detection, motion detection and discrimination, object motion, identification of approaching motion (looming), anticipation, motion extrapolation and omitted stimulus-response \cite{Gollisch2010}. Vertebrate retinas are notable for i) incorporating millions of these retinal circuits, ii) being extremely efficient (the whole human brain consumes approximately 20 Watts) and iii) still displaying the capability to outperform any state-of-the-art computer \cite{TrujilloHerrera2020}. \\

In computer vision, object motion detection is traditionally performed using BS methods, where the foreground (pixels or group of pixels whose light intensity values have suffered an abrupt variation) are compared with the previous image or background model \cite{Piccardi2004,Maddalena2018,garcia2020,Chapel2020}. BS can be static, subtracting the current image frame from the first image frame, or dynamic, subtracting the current image frame from previous image frame(s) \cite{Holmes2018, STALIN2014, Vacavant2013, Sharma2018, Rashid2016, Seo2016}. BS methods can be classified as 1) Mathematical, 2) Machine Learning and 3) Signal processing \cite{Chapel2020,garcia2020}. Mathematical theories are the simplest way to model backgrounds using temporal average, temporal median and histograms, which can be improved using refined models (such as a mixture of Gaussians, kernel density estimation, etc.) and require low computational resources \cite{Chapel2020}. Machine learning models are more robust for performing BS, but they must be trained on the target visual features and require significant computational resources \cite{garcia2020}. Signal processing models used to model the background using the temporal history of pixels as 1D signals and usually require moderate computational resources \cite{Chapel2020}. Although less robust, the classical mathematical BS models are better suited for real-time on near real-time applications. As real-time processing is a key objective of this work, we focus only on mathematical models in this paper. \\
The Hybrid Sensitive Motion Detection (HSMD) model reported in this paper was inspired by the object motion functionality exhibited by vertebrate retinas, in which object motion-sensitive retinal cells (OMS-RC) determine the difference between a local patch's motion trajectory and the background \cite{Gollisch2010}. An improved version of DBS using Local Singular Value Decomposition (SVD) Binary Pattern (mathematical model) \cite{Samsonov2017,samsonovcode2017} is enhanced by a 3-layer spiking neural network (SNN), forming a hybrid architecture. \\

The main contributions of the work reported in this paper are i) an object motion detection model inspired by the OMS-RC designed to work with commercial-of-the-shelf (COTS) cameras, ii) enhancement of the dynamic BS (mathematical model) using the 3-layered SNN and iii) optimisation of the proposed method for processing live capture feeds in near real-time. The algorithm  was tested on the 2012 change detection (CDnet2012) \cite{Goyette2012} and 2014 change detection (CDnet2014) benchmark datasets \cite{Wang2014} and compared with classical BS algorithms (discussed in sections \ref{ch2:lit_review} and \ref{ch4:methodology}). The HSMD  can detect motion using commercial-off-the-shelf camera feeds and/or video clips using Spiking Neural Networks (SNN), as opposed to cameras exploiting dedicated custom architectures.\\
The remainder of the paper is structured as follows: related work on object detection using classical computer vision and bio-inspired computer vision is briefly reviewed in section \ref{ch2:lit_review}; the HSMD is described in section \ref{ch3:architecture}; the training details, use-case scenarios and HSMD parameterisation are described in section \ref{ch4:methodology}; the results are reported and analysed in section \ref{ch5:results}; and the discussion and future work are presented in section \ref{ch6:conclusion}.

\section{Literature Review} \label{ch2:lit_review}
Reliable and optimised object motion detection in videos and live streams are an essential feature for a wide range of computer vision applications such as object tracking, intrusion detection, collision avoidance, etc. Motion detection is performed by analysing/tracking the variation of light intensities between a set of image frames. \\ 
Camera, background and foreground are three factors that affect the quality of the BS \cite{Kalsotra2019}. 
Current  BS challenges include 
(i) abrupt illumination changes, which impact the pixel intensity values and may increase the number of false positives; 
(ii) dynamic objects, where background object movement may interfere with motion detection of static BS; 
(iii) relative motion, where both the camera and the object move at the same time, creating dynamic backgrounds; 
(iv) challenging weather conditions such as fog, rain, snow, air or turbulence generates errors; 
(v) camouflage, where camouflage regions occur when the foreground and background light intensity pixels are similar; 
(vi) occlusion, when another object or fixed structure obstructs the object of interest; 
(vii) irregular object motion - objects that suddenly increase or decrease in speed; 
(viii) noise, possibly arising from dirty lenses, dust, extremely high/low light intensity, etc. which decrease the quality of the detection; 
(ix) bumps and jitter artefacts which occur when the camera is moved; 
(x) image compression, where lossy compression produces a loss of information, with a consequent reduction in performance.

The OpenCV library \cite{OpenCV2020} is one of the most frequently used computer vision libraries and is the reference library for computer or robot vision researchers. It includes several BS algorithms. Stauffer \& Grimson \cite{Stauffer1999}, and KaewTraKulPong \& Bowden \cite{KaewTraKulPong2002} suggest modelling each pixel as a mixture of Gaussians (MOG) where the Gaussian distributions of the adaptive mixture model are analysed for determining which ones are likely to belong to the background process. All the pixel values that do not fit in the background distributions are considered foreground \cite{Stauffer1999}. Zivkovic \cite{Zivkovic2004} proposes an efficient adaptive algorithm using the Gaussian mixture probability density (MOG2) for enhancing the MOG algorithm. MOG2 selects automatically the number of components per pixel which results in full adaptation to the observed scene. Zivkovic \& Heijden \cite{Zivkovic2006} identified recursive equations for updating the parameters of the MOG, and proposed an enhanced algorithm using K nearest neighbours (MOG-KNN) for the automatic selection of the pixel components. The Gaussian mixture based algorithms (MOG, MOG2 and MOG-KNN) show good robustness when exposed to noise and losses due to image compression but lack sensitivity to intermittent object motion, moving background objects and abrupt illumination changes. In 2016, Sagi Zeevi \cite{CNT-2016} proposed the CNT algorithm, which performed better on the CDnet2014 and targets embedded platforms (e.g. Raspberry PI). The CNT uses minimum pixel stability for a specified period for modelling the background; this can vary from 170 ms (default for swift movements) up to hundreds of seconds (the 60s is the default value) \cite{Zeevi2016}.

Godbehere \textit{et al.} \cite{Godbehere2014} suggested a single-camera statistical segmentation and tracking algorithm (GMG) by combining per-pixel Bayesian segmentation, a bank of Kalman filters and Gale-Shapley matching for the approximation of the solution to the multi-target problem. The proposed GMG algorithm is limited when processing video streams susceptible to camouflage, losses due to image compression and noise.
Guo \textit{et al.} \cite{Guo2016} reported an adaptive BS model enhanced by a local singular value decomposition (SVD) binary pattern (LSBP) for addressing illumination changes. The proposed LSBP algorithm enhances the robustness of the motion detection to illumination changes, shadows and noise. However, it is less effective when processing video streams susceptible to camouflage, losses due to image compression and noise. More recently, in 2017, OpenCV released an improved version of the LSBP algorithm, also known as GSOC \cite{GSOC2020,samsonovcode2017}, developed during the Google Summer of code, which enhances the LSBP algorithm by using colour descriptors and various stabilisation heuristics \cite{Samsonov2017,samsonovcode2017}. The GSOC algorithm demonstrates better performance on the CDnet2012 and CDnet2014 datasets \cite{Samsonov2017,OpenCV2021} when compared to other algorithms available on the OpenCV library. The OpenCV's BS algorithms (i.e. MOG, MOG2, CNT, MOG-KNN, GMG, LSBP and GSOC) were designed for modelling the dynamic background changes (i.e. about two hundred frames are required to train the background model) and classifying all the background outliers as foreground. In this paper, the HSMD algorithm uses GSOC for performing the first stage of BS before enhancement by the SNN. The GSOC algorithm was selected over the other BS algorithms available on the OpenCV library because it is the algorithm that has performed better in the target datasets (i.e. CDnet2012 and CDnet2014) \protect\cite{Samsonov2017,OpenCV2021}.
\newline
Spiking neural networks have also been exploited for object motion detection. Wu \textit{et al.} \cite{Wu2008} proposed a bio-inspired spiking neural network to detect moving objects in a visual image sequence. The SNN was trained to extract the boundaries of moving objects from grey images. Cai \textit{et al.} \cite{Cai2012} expanded this work in \cite{Wu2008} and mimicked the basic functionality of motion detection with axonal delays. These two SNN architectures \cite{Wu2008,Cai2012} were to perform basic detection of moving objects, but neither can process moving objects in real-time. 

Lichsteiner \textit{et al.} \cite{Lichtsteiner2005} introduced the concept of an Address-Event Representation (AER) silicon retina chip capable of generating events proportional to the log intensity changes. Farian \textit{et al.} \cite{Farian2015} proposed an in-pixel colour processing approach inspired by the retinal colour opponency, using the same AER concept. Brandli \textit{et al.} \cite{Brandli2014} proposed a new version of the AER camera reported by Lichsteiner \cite{Lichtsteiner2005} called the dynamic, active pixel vision sensor (DAVIS), which exploits the efficiency of the AER protocol and introduces a new synchronous global shutter frame concurrently. In this current paper, the authors used the HSMD algorithm in conjunction with a DAVIS 240C camera and a standard RGB device, and compared the resultant performance in an object tracking scenario.

Kasabov et al. \cite{Kasabov2013} proposed the deSNN that combines the use of Spike Driven Synaptic Plasticity (an unsupervised learning method for learning spatio-temporal representations) with rank-order learning (supervised learning for building rank-order models). The deSNN was tested on data collected by an Address-Event Representation (AER) silicon retina chip \cite{Lichtsteiner2005} (which generates spiking events in response to changes in light intensity) for recognising moving objects. Although able to recognise moving objects, the deSNN was designed to work specifically with AER cameras. Unlike the deSNN, the HSMD has been designed to work with commercial-off-the-shelf RGB cameras. More recently, Jiang \textit{et al.} \cite{Jiang2019} proposed an SNN based on the Hough Transform to detect a target object with an asynchronous event stream fed by an AER camera. The algorithm \cite{Jiang2019} was able to process up to 40.74 frames per second on an Intel i7-4770 processor, accelerated by an Nvidia Geforce GTX 645. However, it is unclear whether the algorithm would work with regular commercial-of-the-shelf (COTS) cameras.\\

Similar to AER silicon retina chip, the HSMD algorithm mimics the basic functionality of OMS-GC with the difference that HSMD works with any COTS camera. Moreover, the HSMD uses a 3-layer SNN to enhance the GSOC algorithm, which has the best results when tested against CDnet2012 and CDnet2014 datasets. To the best of the authors' knowledge, the HSMD is the first SNN-based algorithm capable of processing image streams in near real-time(i.e. 720$\times$480@13.82fps [CDnet2014] and 720$\times$480@13.92fps [CDnet2012]) as a consequence of the parallel optimisations performed in terms of making use of the 96 hyper-threaded cores available the Intel(R) Xeon(R) Platinum 8160 CPU @ 2.10GHz that was used in this work.

\section{HSMD architecture} \label{ch3:architecture}
\begin{figure*}[hbt!]
\centering
\includegraphics[width=1.0\textwidth]{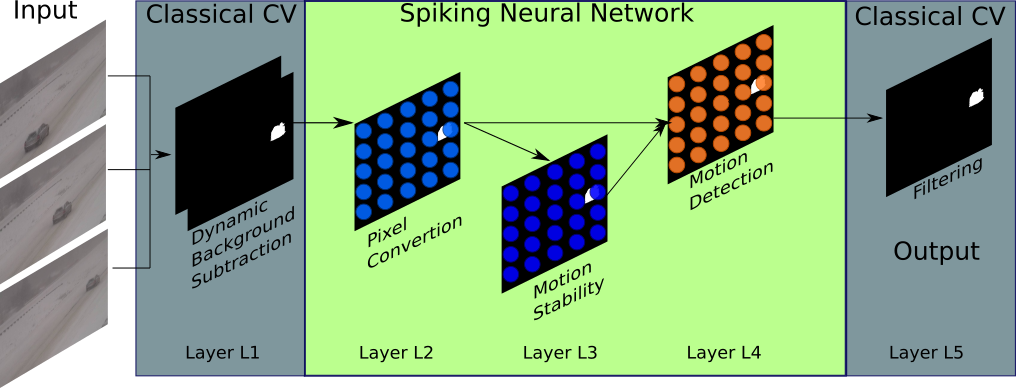}
\caption{HSMD with (i) $n\times m$ image input followed by DBS, three spiking neuronal layers and filtering. Layer 1: DBS, Layer 2: pixel intensity to spike events encoding, Layer 3: Motion stability, Layer 4: motion detection and Layer 5: filtering.
\label{fig:proposed_architecture}} 
\end{figure*}

The HSMD is a combined BS/SNN Network to create a hybrid model for detecting motion, emulating the elementary functionalities of the object-motion-sensitive ganglion cells (OMS-GC) as described in \cite{Gollisch2010}. \newline 
 The architecture of the HSMD is shown in Figure~\ref{fig:proposed_architecture}. There are five layers to the overall architecture. Layer 1 performs the DBS using the GSOC algorithm (described in section \ref{ch4:methodology}). The resulting DBS frames are fed into Layer 2 of the Spiking Neural Network (SNN), where the pixel intensity values are converted into currents that are proportional to the light intensity (see \ref{layer_2}). The DBS-converted currents are fed to the Layer 2 neurons via a 1:1 synaptic connectivity. Layer 2 neurons are synaptically connected to Layer 3 neurons, which performs the first stage of motion analysis; Layer 3 neurons connect to the Layer 4 neurons via 1:1 synaptic connectivity. Layer 4 neurons perform precise motion detection.  A median filter filters their spikes to exclude random neuron activities. 
 
\subsection{Spiking Neuron Models} \label{ch3.1:neurons}
The Leaky-Integrate-and-Fire (LIF) was the spiking neuron model used in this work because of its simplicity, computationally efficiency and suitability for processing images in near real-time. The LIF spiking neuron model exhibits similar, but less complex, dynamics compared to real biological neurons \cite{Gerstner2002}. More complex spiking neuron models are available, e.g. Hodgkin-Huxley, but require significant computational resources and have a higher impact on the computational performance (e.g. Izhikevich \cite{Izhikevich2004}). The LIF neuron's dynamics are described by equation \ref{eq:1}.
\begin{eqnarray}
\label{eq:1}
\tau_m\frac{\delta V_m}{\delta t} =- V_m + RI \left( t \right)\\
\nonumber
\end{eqnarray}
\vspace{-4mm}

\noindent where $\tau_m = RC$ is the time constant, $R$ the membrane resistance, $C$ the membrane capacitance, $V_m(t)$ the membrane voltage and $I(t)$ is the current at time $t$.
The membrane potential $V_m$ is reset to the reset membrane potential ($E_L$) and a spike event is generated when $V_m(t)$ crosses the $V_{th}$ (threshold voltage).

\subsection{Input Layer: DBS and reduction}\label{input_layer}

Each $n \times m$ image frame (i.e. camera, video sequence or image sequences) is converted into grayscale.\\

The GSOC \cite{GSOC2020} delivers an adaptive DBS using colour descriptors and various stabilisation heuristics \cite{Samsonov2017,samsonovcode2017} while processing the frames pixel-wise and leveraging the parallelism inside OpenCV \cite{Samsonov2017}. 

\subsection{Layer 2: Pixel intensities to currents encoding}\label{layer_2}
Pixel intensity values are converted into proportional currents and fed into the spiking neurons in Layer 2 via a 1:1 connectivity. The Layer 1 neurons were trained to trigger spike events proportional to the pixel intensity values, as described by equation~\ref{eq:pixel_current}. 
\begin{equation}
    i_c(x,y)=I(x,y). c
    \label{eq:pixel_current}
\end{equation}

\noindent where $i_c(x,y)$ is the corresponding current for the image light intensity value $I(x,y)$ at coordinates $x$ and $y$, and $c$ is a conversion constant obtained experimentally (in our case, $c$=17.5).

\subsection{Layer 3: Motion stability} \label{layer_3}
Layer 3 is used to perform motion stabilisation through the creation of local buffers by delaying the propagation of spike events. A delay is created when a given neuron of layer 2 connects to a neuron in layer 3, before being passed to Layer 4, instead of the direct Layer 2 to Layer 4 connection. Spike events passing through Layer 3 are buffered by neurons in Layer 3 for one simulation time-step ($\delta t$, in this work $\delta t=10$ ms) and presented to the neurons in Layer 4. N[n] in the following simulation time-step. 

\begin{figure}[H]
\centering
\includegraphics[width=0.45\textwidth]{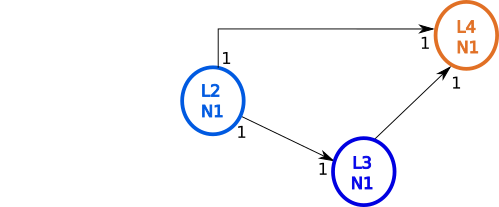}
\caption{HSMD connectivity. In this example, it can be seen that the neuron 1 (N1) of each layer connects to the N1 of the subsequent layer.
\label{fig:connectivity}} 
\end{figure}

The neurons in Layer 2 are connected to the Layer 3 neurons via a 1:1 connectivity. Finally, the Layer 3 neurons connect to the Layer 4 neurons via a 1:1 connectivity as shown in Figure~\ref{fig:connectivity}. All synaptic weights from Layer 2 to Layers 3 and 4 have a value of 1370 (obtained experimentally).

\subsection{Layer 4: Motion detection} \label{layer_4}
The Layer 4 neurons receive synaptic connections from the neurons in Layer 2 and Layer 3 via excitatory synapses and exploit these spiking events to detect motion. Spike events generated by Layer 4 neurons result from dynamic changes between sequential image frames. Signals received directly from Layer 2 neurons enable detection of changes between the current image frame n and the previous image frame n-1. In contrast, those routed via Layer 3 neurons compare the image frame n-1 with the image frame n-2. Layer 4 spike events are mapped into the corresponding area in the original image captured from the camera. The synaptic weights obtained experimentally are 1370 for all the synapses. The Layer 2 to Layer 4 weights were tuned to forward all the spike events generated in Layer 2. The Layer 3 to Layer 4 synaptic weights were tuned to produce spike events from the Layer 4 neurons for each group of two sequential spike events. The main goal is to give high importance (larger weight) to new spike events (frame [n] - frame [n-1]) and to give lower importance to older spike events (frame [n-1] - frame [n-2]).

\subsection{Layer 5: Filtering} \label{layer_5}
The Layer 4 neurons' spike events matrix is mapped into a motion matrix $M_d$ of the same size as the captured image (i.e. $n\times m$). The events in the $M_d$ matrix are filtered using an averaging filter described by equations \ref{eq:filter} and ~\ref{eq:conv}:

\begin{equation}
H(u,v)=\frac{1}{u.v}\begin{pmatrix}
\begin{bmatrix}
w_{0,\,0} & ... & w_{0,\,u} \\ 
 ... & ...  & ... \\ 
w_{v,\,0} & ... & w_{v,\,u}
\end{bmatrix}
\end{pmatrix}
\label{eq:filter}
\end{equation}

\begin{equation}
Y_d(x,y)=M_d(x,y)*H(u,v)
\label{eq:conv}
\end{equation}

\noindent where $Y_d(x,y)$ is the filtered motion detection matrix, $H(u,v)$ is the averaging filter, $u$ and $v$ are the convolution window length and height respectively, $*$ is the convolution operator, $w$ is the filter window.

\section{Methodology} \label{ch4:methodology}

The HSMD was implemented in C++ using the C++ Standard Template Library 17 (C++17) \cite{StandardC++2017} (implementation of data structures), Boost 1.71 \cite{Boost2020} (file management) and OpenCV 4.5.0 \cite{OpenCV2020} (which provides common computer vision functionalities such as resize, capture and display images) in Ubuntu 20.04 LTS\footnote{Available online \protect\url{http://releases.ubuntu.com/20.04/}, last accessed 12/11/2020}. 

\subsection{HSMD setup} \label{ch4.2:setup}
The HSMD initial setup includes the following steps:

\textbf{Step 1 - Select between live capture, video analysis or image sequences:} The user can opt to run the algorithm directly on images being captured by the camera or provide the path of a video or set of image sequences for motion analysis.

\textbf{Step 2 - Create the Layer 2 to Layer 4 neural network:} Read the first image and compute the size of the image. The number of neurons is computed automatically from the dimensions of the first image in a sequence of images.

\textbf{Step 3 - Set the neuronal parameters:} The LIF parameters recommended in the references \cite{Jolivet2004, Brette2005} and frequently used in LIF SNN circuits, were used to configure the SNN. Therefore, the simulation was configured with a time step of $ \delta t$=10 ms and the default neuron parameters as follows: initial $V_m$=-55.0 mV, $E_L$ = -55.0 mV, $C_m$ = 10.0 pF, $R_m$=1.0 MOhm, $V_{reset}$=-70.0 mV, $V_{min}$=-70.0 ms, $V_{th}$=-70.0 mV, $\tau_m$=10.0 ms, $t_{ref}$=2 ms, $w_{syn}$ = 1555.0 (neurons L3 and L4) and $w_{p2i}$=8.0 (L2 neurons only). 

\textbf{Step 4 - Start the image acquisition :} Images are collected from devices, video streams or obtained from folders with sequences of images while the HSMD algorithm is being executed. The pseudo-code of the main algorithm is described in Algorithm~\ref{alg:main}.

\begin{algorithm}
     \caption{HSMD main algorithm pseudo-code}
     \label{alg:main}
    \begin{algorithmic}[1]
    \STATE $newImage = capture\_image\_camera\;$
    \STATE $newImageGrey=colour2grey(newImage)\;$
    \STATE $set\_number\_neurons\_from\_newImageGrey\_shape\;$
    \STATE $build\_neuronal\_network\;$
    \STATE $load\_pretrained\_weights\;$
    \WHILE{frames available}
        \STATE $reset\_spike\_events\;$
        \STATE $newImage = capture\_image\_camera\;$
        \STATE $newImageGrey=colour2grey(newImage)\;$
        \STATE $newImageReduced=newImageGrey\;$
        \STATE $dynSubImage=newImageReduced-previousImage\;$
        \STATE $ previousImage=newImageReduced\;$
        \FOR{I\ in\ dynSubImage}
            \IF{dynSubImage[I]$<$ Threshold}
                \STATE $dynSubImage[I]=0.0\;$
            \ENDIF
            \STATE $currents=convPixel2Current(dynSubImage)\;$
            \FOR{i:=0 \TO timestep}
                \STATE $apply\_currents\_to\_neurons\_L2\;$
                \STATE $compute\_L2\_neuron\_spikes;$
                \STATE $convert\_L2\_neuron\_spikes\_to\_currents;$
                \STATE  $compute\_L3\_neuron\_spikes;$
                \STATE $convert\_L3\_neuron\_spikes\_to\_currents;$
                \STATE $compute\_L4\_neuron\_spikes;$
            \ENDFOR
            \STATE $spikes=get\_sumSpikeEventsPerL4Neuron()\;$
            \STATE $masked\_spikes=applyAveragingFilter(spikes)\;$
            \STATE $spikes=normalise(spikes)\;$
            \STATE $display(newImage)\;$
            \STATE $display(spikes)\;$
       \ENDFOR
    \ENDWHILE
  \STATE $Display\_spike\_rates\;$
	\end{algorithmic}
\end{algorithm}

\subsection{Datasets and metrics} \label{ch4.3:datasets_metrics}
\subsubsection{Datasets} \label{ch4.3.1:datasets}
The CDnet2012 \cite{Goyette2012} (cited more than 379 times\footnote{Retrieved from, \protect\url{https://ieeexplore.ieee.org/abstract/document/6238919}, last accessed: 12/10/2020}) and CDnet2014 \cite{Wang2014} (cited more than 300 times\footnote{Retrieved from, \protect\url{https://ieeexplore.ieee.org/document/6910011}, last accessed: 12/10/2020}) benchmark datasets were designed for benchmarking BS algorithms. While the  HSMD algorithm has been designed as an object detection algorithm and not a BS algorithm, nevertheless these two datasets provide challenging scenarios for robust comparable assessment of the proposed algorithm and network. Within the two benchmark datasets tyhe HSMD was compared with the following state-of-the-art BS algorithms available on the OpenCV library: MOG~ \cite{Stauffer1999}, MOG2~\cite{Zivkovic2004}, MOG-KNN~\cite{Zivkovic2006}, GMG~\cite{Godbehere2014}, LSBP~\cite{Guo2016}, CNT~\cite{CNT-2016} and GSOC~\cite{samsonovcode2017} methods. The OpenCV BS algorithms were used because they are highly optimised, reliable, and publicly available to anyone who wants to test or compare their algorithms.

Each of the benchmark videos in the CDnet2012 \cite{Goyette2012} and CDnet2014 \cite{Wang2014} are considered under one or more challenge categories as follows:\\
\vspace{.1cm}

\begin{figure}[h]
\centering
\includegraphics[width=0.48\textwidth]{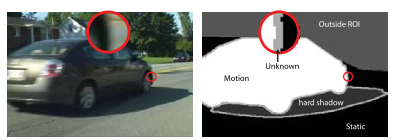}
\caption{Raw image frame (left) and its respective ground-truth (right). The ground-truth images show the annotations using the datasets labels. Adapted from \protect\cite{Goyette2012}}. \label{fig:gt}
\end{figure}

\textbf{CDnet2012 and CDnet2014}
\begin{itemize}
  \item \textbf{Baseline} - reference videos which are relatively simple to classify; some videos contain very simple movements from the next four categories.
  \item \textbf{Dynamic Background} - videos that have both foreground and background motion (e.g. water movement and shaking trees).
  \item \textbf{Camera Jitter} - videos captured with cameras installed on unstable structures.
  \item \textbf{Shadow} - videos containing narrow shadows from solid structures or moving objects.
  \item \textbf{Intermittent Object Motion} videos that include objects that are static for most of the time and suddenly start moving.
  \item \textbf{Thermal} - videos that exhibit thermal artefacts (i.e. bright spots and thermal reflections on windows and floors).
  \end{itemize}
 \textbf{CDnet2014 only}
\begin{itemize}
  \item \textbf{Challenging Weather}: Outdoor videos showing very-low visibility winter storm conditions.
  \item \textbf{Low Frame-Rate}: videos capture at varying frame rates between 0.17 and 1 fps;
  \item \textbf{Night}: includes traffic videos with low visibility and strong headlights. 
  \item \textbf{Pan, Tilt and Zoom (PTZ)}: videos recorded with cameras exposed to PTZ movements. 
  \item \textbf{Air Turbulence}: videos filmed from distances of 5 to 15 km exhibiting air turbulence and frames distortion.
\end{itemize}
The BS algorithms were configured with the default OpenCV settings \cite{OpenCV2020} and compared against the HSMD algorithm.The ground-truth provided by the datasets is composed of the following labels \cite{Goyette2012,Wang2014}:
\begin{itemize}
    \item \textbf{Static} - grayscale value 0;
    \item \textbf{Shadow} - grayscale value 50;
    \item \textbf{non-Region of Interest (RoI)} - grayscale value 85;
    \item \textbf{Unknown} - grayscale value 170;
    \item \textbf{Moving} - grayscale value 255;
\end{itemize}
The \textit{static} and \textit{moving} classes contain pixels that belong to the background and foreground, respectively; The \textit{shadows}, one of the most challenging artefacts, should be classified as part of the background; The \textit{unknown} region should not be considered either background or foreground because it contains pixels that cannot be accurately classified as background or foreground. The non-ROI pixels serve to exclude frames from being classified because some BS algorithms require several pixels for the model to stabilise (i.e. create the background model) and for preventing corruption by non-related activities to the considered category \cite{Goyette2012,Wang2014}. Figure~\ref{fig:gt} shows the 5 class regions.
\subsubsection{Metrics} \label{ch4.3.2:metrics}
The average performance obtained for each category using each BS method and the HSMD algorithms is characterised via eight metrics, as shown below. $TP$ is the number of true positives, $TN$ the number of true negatives, $FN$ the number of false negatives, and $FP$ the number of false positives \cite{Goyette2012,Wang2014}.
\newline
\newline

\begin{enumerate}
\item Recall ($Re$): $R_e=\frac{TP}{TP+FN}$;\\
$Re$: rank by \textbf{descending order} order;\\
\item Specificity ($Sp$): $Sp=\frac{TN}{TN+FP}$;\\
$Sp$ rank by \textbf{descending order} order;\\
\item False Positive Rate (FPR): $FPR=\frac{FP}{FP+TN}$;\\
$FPR$ rank by \textbf{ascending order} order;\\
\item False Negative Rate (FNR): $FNR=\frac{FN}{FN+TP}$;\\
$FNR$ rank by \textbf{ascending order} order;\\
\item Wrong Classifications Rate (WCR):\\ $WCR=\frac{FN+FP}{TP+FN+FP+TN}$;\\
$WCR$ rank by \textbf{ascending order} order;\\
\item Correct Classifications Rate (CCR):\\ $CCR=\frac{TP+TN}{TP+FN+FP+TN}$;\\
$CCR$ rank by \textbf{descending order} order;\\
\item Precision (Pr): $Pr=\frac{TP}{TP+FP}$;\\
$Pr$ rank by \textbf{descending order} order;\\
\item F-measure (F1): $F1=2\times \frac{PR.RE}{PR+RE}$\\
$F1$ rank by \textbf{descending} order;\\
\end{enumerate}
These eight metrics contribute to the overall average ranking ($R$) and overall average ranking across all categories ($\overline{RC}$).\\ 
\\
Average Ranking (R):
$R=\frac{\overline{Re}+\overline{Sp}+\overline{FPR}+\overline{FNR}+\overline{WCR}+\overline{CCR}+\overline{F1}}{nMet}$;\\
$R$ rank by \textbf{ascending} order;\\
\\
Average Ranking across all Categories ($\overline{RC}$):\\ 
$\overline{RC}=\frac{RE+SP+FPR+FNR+WCR+CCR+F1}{nMet}$;\\
$\overline{RC}$ rank by \textbf{ascending} order;\\
\\
where $nMet$ is the number of metrics (8 in this case). 

\section{Results} \label{ch5:results}
The HSMD was tested on both datasets under the same conditions to ensure an accurate and rigorous comparison. The results are presented both as overall results and per category to understand better the specific performances obtained per method. The overall results for each method are presented in section \ref{ch5.1:overall results} and the results per method and category in section \ref{ch5.2:results_category}.

Please note that in the tables (\ref{tab:results_overall_12} to \ref{tab:results_category}), the $\uparrow$ means that the highest score is the best result and the $\downarrow$ that the lowest result is the best result. The best results are highlighted using light grey for all the methods except the HSMD results highlighted in dark grey. $Re$ is the Recall, $Sp$ is the Specificity, FPR is the False Positive Rate, FNR is the False Negative Rate, WCR is the Wrong Classifications Rate, CCR is the Correct Classifications Rate, Pr is the Precision, F1 is the F score or F-measure, R is the Average Ranking and $\overline{RC}$ is the Average Ranking across all Categories.

\begin{figure*}[htb!]
\centering
\includegraphics[width=0.96\textwidth]{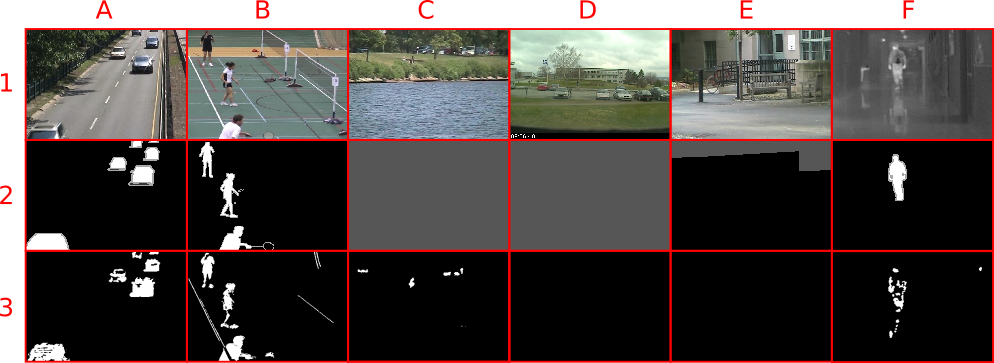}
\caption{Results obtained for each of the five categories common to both CDnet2012 and CDnet2014 datasets. Column A: baseline; B: camera jitter, C: dynamic background; D: dynamic object motion; E: shadow and F: thermal. Row 1: RGB image; 2: ground-truth; and 3: HSMD binarised. The raw images, shown in the first row, demonstrate the scenarios that can be found in both datasets. The corresponding ground truth images, presented in the second row, show the 5 labels, namely, i) static [grayscale value 0], ii) shadow [grayscale value 50], iii) non-ROI [grayscale value 85], iv) unknown [grayscale value 170] and v) moving [grayscale value 255]. The corresponding binarised images generated by the HSMD are shown in the third row.} \label{fig:CDnet2012_output}
\end{figure*}

Figure~\ref{fig:CDnet2012_output} shows the results obtained for each of the five categories common to both CDnet2012 and CDnet2014.

\subsection{Overall results} \label{ch5.1:overall results}
Tables~\ref{tab:results_overall_12} and \ref{tab:results_overall_14} present the overall results obtained per method and per metric, ranked by $\overline{RC}$ (average ranking across all categories, first column) in ascendant order.

\begin{table*}[t]\caption{CDnet2012 overall results\strut} \label{tab:results_overall_12}
\resizebox{17.5cm}{!}{
\begin{tabular}{|l|>{\columncolor[gray]{0.8}}l|l|l|l|l|l|l|l|l|l|}
\hline
\rowcolor[gray]{0.75}
Method & $\overline{RC}$ $\downarrow$ & Re $\uparrow$ & Sp $\uparrow$ & FPR $\downarrow$ & FNR $\downarrow$ & WCR $\downarrow$ & CCR $\uparrow$ & F1 $\uparrow$ & Pr $\uparrow$ \\ \hline\hline
\cellcolor{gray}\textcolor{white}{\protect\textbf{HSMD}}&\cellcolor{gray}\textcolor{white}{\protect\textbf{2.8}}&0.52&0.994&0.006&0.23&\cellcolor{gray}\textcolor{white}{\protect\textbf{0.024}}&\cellcolor{gray}\textcolor{white}{\protect\textbf{0.976}}&\cellcolor{gray}\textcolor{white}{\protect\textbf{0.77}}&0.62\\\hline
GSOC&3.5&0.54&0.993&0.007&0.25&\cellcolor{gray!20}\textit{0.024}&\cellcolor{gray!20}\textit{0.976}&0.75&\cellcolor{gray!20}\textit{0.63}\\\hline
MOG2&3.8&0.37&0.995&0.004&0.24&0.026&0.974&0.76&0.50\\\hline
GMG&3.9&0.20&\cellcolor{gray!20}\textit{0.998}&\cellcolor{gray!20}\textit{0.002}&\cellcolor{gray!20}\textit{0.21}&0.033&0.967&0.79&0.32\\\hline
KNN&4.3&0.39&0.995&0.005&0.26&0.025&0.975&0.74&0.51\\\hline
MOG&4.5&0.32&0.996&0.004&0.26&0.030&0.970&0.74&0.44\\\hline
CNT&6.1&\cellcolor{gray!20}\textit{0.73}&0.927&0.073&0.71&0.081&0.919&0.29&0.41\\\hline
LSBP&7.3&0.57&0.90&0.096&0.80&0.109&0.891&0.20&0.29\\\hline
\end{tabular}}
\\
$\uparrow$: the highest score is the best.\\
$\downarrow$: the lowest result is the best. \\
Best HSMD results are highlighted using dark grey, while best results per category are highlighted with light grey for other methods.
$Re$ is the Recall, $Sp$ is the Specificity, FPR is the False Positive Rate, FNR is the False Negative Rate, WCR is the Wrong Classifications Rate, CCR is the Correct Classifications Rate, Pr is the Precision, F1 is the F score or F-measure and $\overline{RC}$ is the Average Ranking across all Categories.
\end{table*}

From Table~\ref{tab:results_overall_12} (2nd column), it may be seen that the HSMD algorithm ranks in first place across all eight methods with which it is compared when tested on the CDnet2012 dataset. Although the HSMD performed very well in 5 of the eight metrics, it is essential to highlight the WCR, CCR and F1 metrics results. The results show that the HSMD is sensitive to object motion due to the highest correct counts and lowest wrong counts rate, which contribute to getting the highest F-score and the second-best Precision. Furthermore, it is possible to conclude that the HSMD has improved the performance of the GSOC algorithm compared to when it is used alone when tested on the six categories of the CDnet2012.

The HSMD has also performed very well when tested on the CDnet2014 (see Table~\ref{tab:results_overall_14}).

\begin{table*}[htb!]\caption{CDnet2014 overall results\strut} \label{tab:results_overall_14}
\resizebox{17.5cm}{!}{\begin{tabular}{|l|>{\columncolor[gray]{0.8}}l|l|l|l|l|l|l|l|l|l|}
\hline
\rowcolor[gray]{0.75}
Method & $\overline{RC}$ $\downarrow$ & Re $\uparrow$ & Sp $\uparrow$ & FPR $\downarrow$ & FNR $\downarrow$ & WCR $\downarrow$ & CCR $\uparrow$ & F1 $\uparrow$ & Pr $\uparrow$ \\ \hline\hline
\cellcolor{gray}\textcolor{white}{\protect\textbf{HSMD}}&\cellcolor{gray}\textcolor{white}{\protect\textbf{2.9}}&0.55&0.993&0.007&0.35&0.018&0.982&0.65&\cellcolor{gray}\textcolor{white}{\protect\textbf{0.60}}\\\hline
GSOC&3.0&0.40&0.995&0.005&0.38&\cellcolor{gray!20}\textit{0.017}&\cellcolor{gray!20}\textit{0.983}&0.62&0.48\\\hline
KNN&3.5&0.34&0.996&0.004&\cellcolor{gray!20}\textit{0.32}&0.019&0.981&\cellcolor{gray!20}\textit{0.68}&0.45\\\hline
GMG&4.3&0.24&\cellcolor{gray!20}\textit{0.997}&\cellcolor{gray!20}\textit{0.003}&0.36&0.022&0.978&0.64&0.35\\\hline
MOG&4.4&0.58&0.991&0.009&0.39&0.019&0.981&0.61&0.60\\\hline
MOG2&4.5&0.39&0.994&0.006&0.42&0.018&0.982&0.58&0.47\\\hline
LSBP&6.5&0.58&0.945&0.055&0.79&0.064&0.936&0.21&0.31\\\hline
CNT&7.0&\cellcolor{gray!20}\textit{0.72}&0.930&0.070&0.80&0.075&0.925&0.20&0.32\\\hline
\end{tabular}}
$\uparrow$: the highest score is the best result.\\
$\downarrow$: the lowest result is the best. \\
Best results per category are highlighted using dark grey for the HSMD and light grey for other methods.
Where $Re$ is the Recall, $Sp$ is the Specificity, FPR is the False Positive Rate, FNR is the False Negative Rate, WCR is the Wrong Classifications Rate, CCR is the Correct Classifications Rate, Pr is the Precision, F1 is the F score or F-measure and $\overline{RC}$ is the Average Ranking across all Categories.
\end{table*}

Table~\ref{tab:results_overall_14} shows that the HSMD algorithm was ranked in first place in the Average Ranking across all Categories column when tested on the CDnet2014 dataset. The HSMD performed very well in 7 of the eight metrics and exceptionally well on the Precision metrics. It can also be seen that there was a slight decrease in the HSMD performance when tested on the eleven categories available on the CDnet2014 as compared to the original six of the CDNet2012 dataset. The result is to be expected; none of the methods has excellent performance across all metrics.

\subsection{Results obtained per category} \label{ch5.2:results_category}

The Average Ranking (R) for each of the methods per category is shown in Table~\ref{tab:results_category}.

\begin{table*}[]
\begin{center}
\caption{Results per category} \label{tab:results_category}
\resizebox{17.5cm}{!}{
\begin{tabular}{|l|l|l|l|l|l|l|l|l|l|l|l|l|l|l|l|l|l|l|l|l|l|}
\hline
\rowcolor[gray]{0.75}
\multicolumn{2}{|c|}{IntObjMotion}&
\multicolumn{2}{|c|}{shadow} & 
\multicolumn{2}{|c|}{cameraJitter} &
\multicolumn{2}{|c|}{badWeather} &
\multicolumn{2}{|c|}{dynamicBackground} &
\multicolumn{2}{|c|}{nightVideos} &
\multicolumn{2}{|c|}{PTZ}&
\multicolumn{2}{|c|}{thermal} & 
\multicolumn{2}{|c|}{baseline} & 
\multicolumn{2}{|c|}{lowFramerate} & 
\multicolumn{2}{|c|}{turbulence} 
\\\hline
\rowcolor[gray]{0.75}
Method & R$\downarrow$ & 
Method & R$\downarrow$ & 
Method & R$\downarrow$ &
Method & R$\downarrow$ &
Method & R$\downarrow$ &
Method & R$\downarrow$ &
Method & R$\downarrow$ & 
Method & R$\downarrow$ & 
Method & R$\downarrow$ & 
Method & R$\downarrow$ & 
Method & R$\downarrow$ 
\\\hline\hline
\cellcolor{gray}\textcolor{white}{\protect\textbf{HSMD}}&
\cellcolor{gray}\textcolor{white}{\protect\textbf{3.875}}&
\cellcolor{gray!20}\textit{MOG2}&
\cellcolor{gray!20}\textit{2.375}&
\cellcolor{gray!20}\textit{LSBP}&
\cellcolor{gray!20}\textit{1.875}&
\cellcolor{gray!20}\textit{CNT}&
\cellcolor{gray!20}\textit{2.125}&
\cellcolor{gray!20}\textit{LSBP}&
\cellcolor{gray!20}\textit{1.75}&
\cellcolor{gray}\textcolor{white}{\protect\textbf{HSMD}}&
\cellcolor{gray}\textcolor{white}{\protect\textbf{2.625}}&
\cellcolor{gray!20}\textit{CNT}&
\cellcolor{gray!20}\textit{1.75}&
\cellcolor{gray}\textcolor{white}{\protect\textbf{HSMD}}&
\cellcolor{gray}\textcolor{white}{\protect\textbf{3.25}}&
\cellcolor{gray!20}\textit{MOG2}&
\cellcolor{gray!20}\textit{2.625}&
\cellcolor{gray!20}\textit{GSOC}&
\cellcolor{gray!20}\textit{2.25}&
\cellcolor{gray}\textcolor{white}{\protect\textbf{HSMD}}&
\cellcolor{gray}\textcolor{white}{\protect\textbf{2.875}} 
\\\hline

LSBP&4.125&CNT&3.5&CNT&2.625&\cellcolor{gray}\textcolor{white}{\protect\textbf{HSMD}}&\cellcolor{gray}\textcolor{white}{\protect\textbf{2.5}}&MOG&2.375&LSBP&3.625& MOG&3.0&MOG&3.375&MOG&3.125&GMG&3.0&MOG2&2.875\\\hline

GMG&4.125&\cellcolor{gray}\textcolor{white}{\protect\textbf{HSMD}}&\cellcolor{gray}\textcolor{white}{\protect\textbf{3.875}}&GSOC&4.125&KNN&3.5&KNN&3.625&MOG2&4.0&LSBP&4.25&GSOC&3.625&\cellcolor{gray}\textcolor{white}{\protect\textbf{HSMD}}&\cellcolor{gray}\textcolor{white}{\protect\textbf{3.625}}&CNT&3.375&LSBP&3.875\\\hline

GSOC&4.5&KNN&4.25&\cellcolor{gray}\textcolor{white}{\protect\textbf{HSMD}}&\cellcolor{gray}\textcolor{white}{\protect\textbf{4.5}}&GMG&3.75&CNT&4.125&GSOC&4.25&MOG2&4.875&KNN&3.75&GMG&4.0&LSBP&4.875&KNN&4.375\\\hline

MOG2&4.875&MOG&4.375&GMG&4.625&GSOC&5.5&MOG2&5.5&MOG&5.25&\cellcolor{gray}\textcolor{white}{\protect\textbf{HSMD}}&\cellcolor{gray}\textcolor{white}{\protect\textbf{4.875}}&CNT&4.75&GSOC&4.875&MOG2&5.125&CNT&4.375\\\hline

MOG&5.0&GMG&5.5&KNN&4.875&LSBP&5.875&GMG&5.875&CNT&5.25&KNN&5.5&LSBP&5.25&KNN&5.875&MOG&5.25&GSOC&5.375\\\hline

KNN&6.375&GSOC&6.0&MOG&6.625&MOG2&6.125&GSOC&6.25&GMG&5.5&GSOC&5.5&GMG&5.25&LSBP&5.875&KNN&5.5&MOG&6.0\\\hline

CNT&6.75&LSBP&6.125&MOG2&6.75&MOG&6.625&\cellcolor{gray}\textcolor{white}{\protect\textbf{HSMD}}&\cellcolor{gray}\textcolor{white}{\protect\textbf{6.5}}&KNN&5.5&GMG&6.25&MOG2&6.75&CNT&6.0&\cellcolor{gray}\textcolor{white}{\protect\textbf{HSMD}}&\cellcolor{gray}\textcolor{white}{\protect\textbf{6.625}}&GMG&6.25\\\hline
\end{tabular}}
\end{center}
$\downarrow$: the lowest result is the best.\\
The HSMD results are highlighted using dark grey for the HSMD, and the best results of other methods are highlighted using light grey.\\
R is the average Ranking.
\end{table*}

\begin{figure*}[]
\centering
\includegraphics[width=1.0\textwidth]{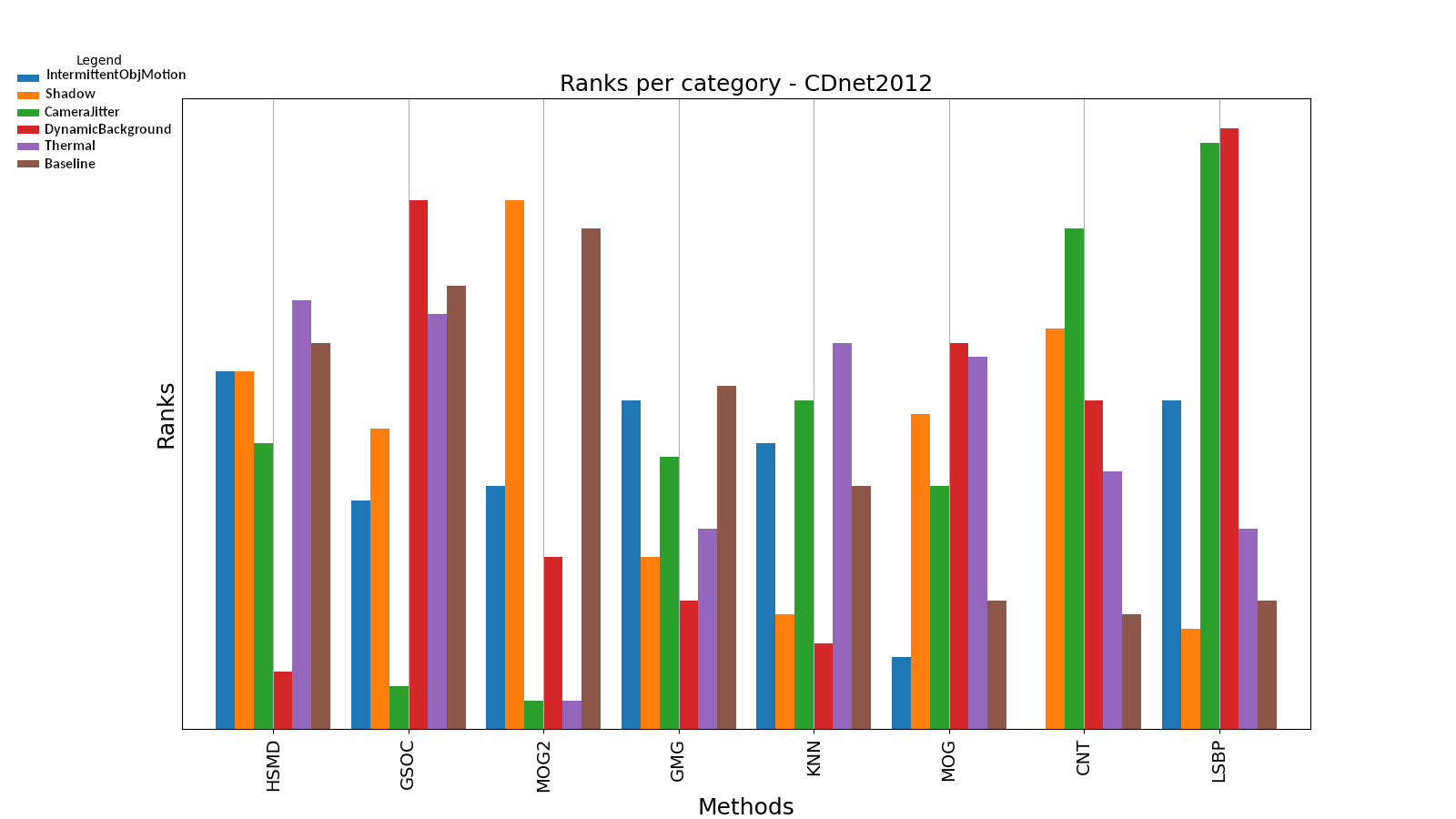}
\caption{CDnet2012 overall results per category and method. The highest bars show the higher ranks, and it is clear that none of the methods had the best ranks in all the categories. Furthermore, it is possible to see that the HSMD achieved high ranks across all the categories with the exception of dynamic background.} \label{fig:CDnet2012_results}
\end{figure*}

\begin{figure*}
\centering
\includegraphics[width=1.0\textwidth]{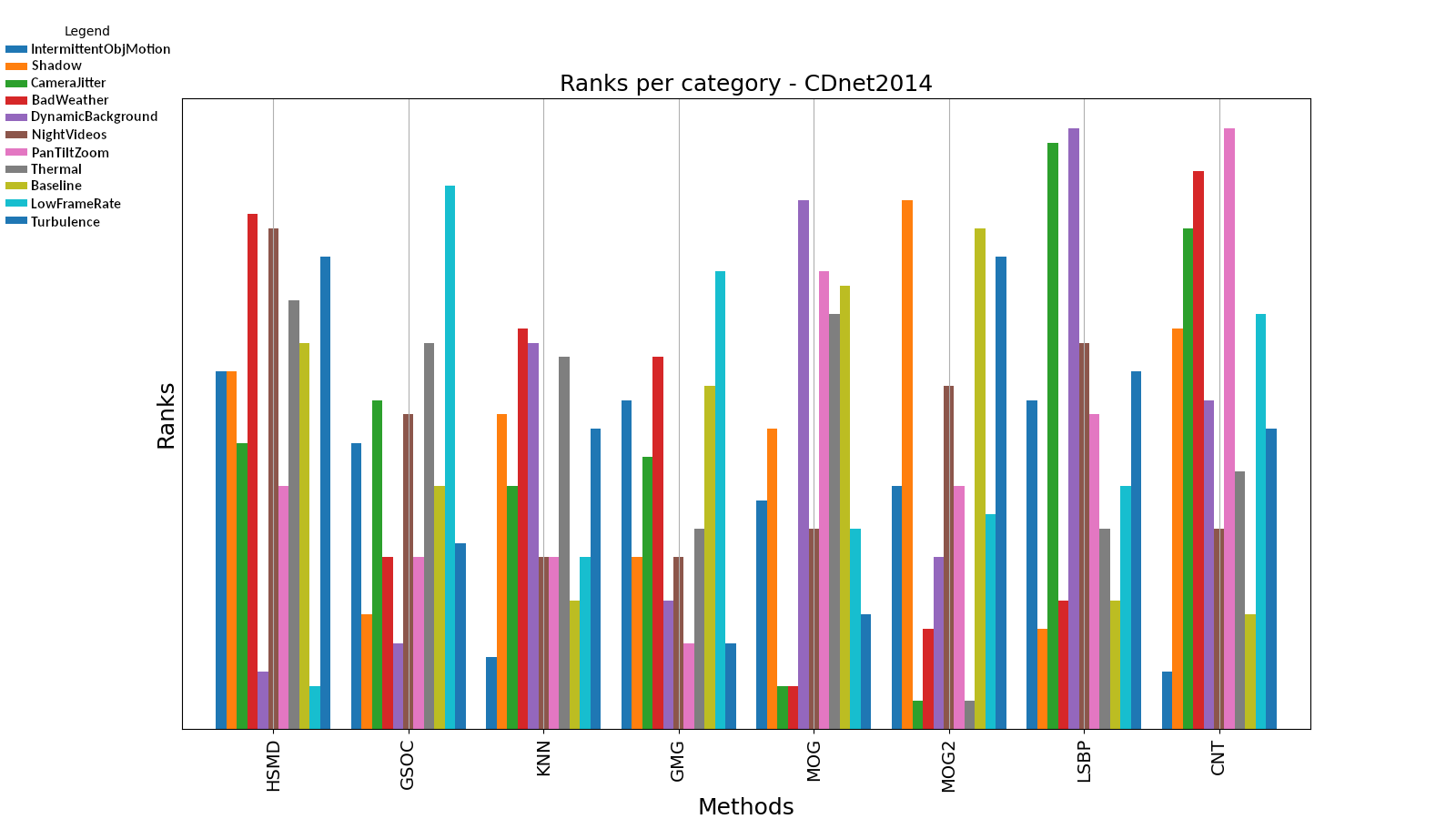}
\caption{CDnet2014 overall results per category and method. The highest bars show the higher ranks, and it is clear that none of the methods had the best ranks in all categories. Furthermore, it is possible to see that the HSMD achieved high ranks across most of the categories except dynamic background and low frame rate.} \label{fig:CDnet2014_results}
\end{figure*}

Figures~\ref{fig:CDnet2012_results}~and~\ref{fig:CDnet2014_results} show the variation of the ranks obtained per category and per method. 

From the analysis of the results shown in Table~\ref{tab:results_category} and Figures~\ref{fig:CDnet2012_results}~and~\ref{fig:CDnet2014_results} it is possible to infer that the HSMD is sensitive to intermittent object motion, night vision, baseline and turbulence; 
These categories share the fact that they relate to moving objects that have high contrast, which is ideal for sensing by the spiking neurons. The HSMD has improved the results of the GSOC in 8 of the 11 categories, except for the low frame rate, dynamic background and camera jitter categories. 
 It is also easy to visualise that the HSMD exhibits the lowest R variation, which justifies why the HSMD was ranked in the first place. 

\subsection{Results analysis}
\label{ch5.1.4:results_analysis}

The HSMD performed very severely in the dynamic background and low frame rate categories, suggesting that the spiking neuron model is not ideal for distinguishing the type of motion. i.e. the spiking neurons detect motion but are unable to distinguish between a shadow or the object itself. This result is probably because, in vertebrate retinas, only the ganglion cells are spiking cells suggesting that distinction between the main object and shadows is probably performed by other non-spiking cells. Nevertheless, the creation of the new approach incorporating both the GSOC algorithm and the SNN, which emulates the basic OMS-GC functionality, clearly improves the accuracy of the GSOC algorithm.\\
The CDnet2012 and CDnet2014 datasets are composed of image files of different resolution, and accordingly, the processing times vary. The HSMD takes approximately 72.4ms (CDnet2014) and 71.9ms (CDnet2012) to process images of 720$\times$480 on a 96-cores Intel(R) Xeon(R) Platinum 8160 CPU @ 2.10GHz equipped with 792 GB of DDR4 and 12.7 TB of disk space. The slight variations are related to other applications running in the background. Therefore, the HSMD is capable of processing images of 720$\times$480 at an average speed of 13.82fps (CDnet2014) and 13.92fps (CDnet2012). Finally, the HSMD is the first hybrid SNN algorithm capable of processing images at such a frame rate, as far as the authors are aware.

\section{Conclusion and Future Work} \label{ch6:conclusion}

A bio-inspired hybrid spiking neural network (HSMD) has been proposed to detect object motion and assess against the CDnet2012 and CDnet2014 datasets. These incorporate video sequences of many moving objects under various challenging environmental conditions and are widely used for benchmarking BS algorithms. The CDnet2012 is composed of 6 categories of movements, and the CDnet2014 augments the initial 6 to 11 categories of movements. Eight metrics, utilised as standard in the CDnet datasets, were used to assess and compare the quality of the HSMD algorithm. The HSMD algorithm performed overall best in both the CDnet2012 and CDnet2014 while performing better than all the tested DBS algorithms  in the intermittent object motion, night videos, thermal and turbulence categories, second best in the bad weather category and the third-best on the baseline and shadow categories. The comparatively good results are a consequence of using the SNN for emulating the basic functionality of OMS-GC, which improves the sensitivity of the HSMD to object motion. The HSMD is also the first hybrid SNN algorithm capable of processing video/image sequences with near real-time performance (i.e. 720$\times$480@13.82fps [CDnet2014] and 720$\times$480@13.92fps [CDnet2012]). \\
Future work includes optimising the HSMD algorithm to detect and track motion in challenging scenarios (e.g. low frame rate, dynamic background and camera jitter) and an investigation to verify if the SNN improves the output for all the remaining methods. Furthermore, the authors will also test if the SNN can enhance other BS algorithms available on the OpenCV library. The authors also aim to design and implement the HSMD approach on Multi-Processor System-on-Chip (MPSoC) technology and high-end Field-Programmable Arrays (FPGA) hardware to accelerate the HSMD algorithm and b) reduce the power consumption for embedded applications.\\

\bibliographystyle{ieeetr}
\bibliography{IEEEabrv,ref}

\begin{IEEEbiography}
[{\includegraphics[width=1in,height=1.25in,clip,keepaspectratio]{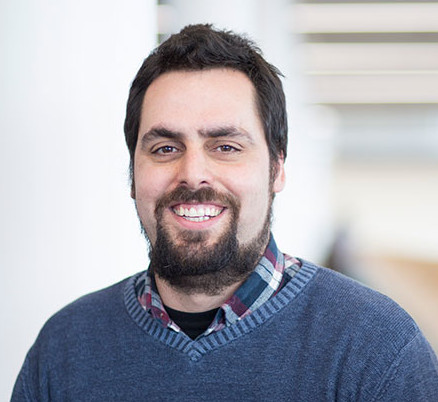}}]{Pedro Machado} received his MSc in Electrical and Computers Engineering from the University of Coimbra (2012) and is currently doing a part-time PhD in Computer Sciences at Nottingham Trent University.
Pedro's expertise includes FPGA design, computer vision, bio-inspired computing, robotics and computational intelligence. His research interests in computer science are retinal cell understanding, biological nervous system modelling, spiking neural networks, robotics and autonomous systems, applied computer vision and neuromorphic hardware.
\end{IEEEbiography} 

\vspace{-1cm}
\begin{IEEEbiography}
[{\includegraphics[width=1in,height=1.25in,clip,keepaspectratio]{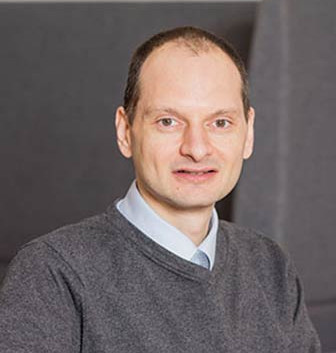}}]{Dr Andreas Oikonomou}
holds a B.Sc. in Engineering (1999), M.Sc. in Information Technology for Management (2000) and Ph.D. in Computer Science with a specialisation in Biomedical Computing (2005). All completed at Coventry University in the UK. Dr Oikonomou has previously taught Computer Science and Games Development at Derby and Coventry Universities in the UK and has also worked as a Project Manager, Quality Assurance Manager and Games Studio head in the software development industry. He has +40 publications on subjects including Image Processing, Interaction Design and Software Simulations. His current research focuses on Artificial Intelligence and Robotics. He is also a member of the British Computer Society.
\end{IEEEbiography} 

\vspace{-1cm}
\begin{IEEEbiography}
[{\includegraphics[width=1in,height=1.25in,clip,keepaspectratio]{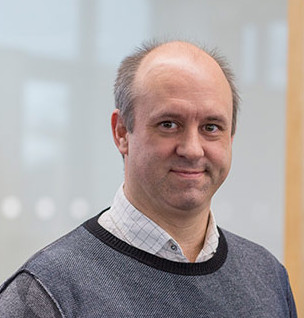}}]{Dr Jo\~ao Filipe Ferreira} (MIEEE, MBCS)
received Ph.D., M.Sc. and B. Sc. degrees in Electrical Engineering from the Faculty of Sciences and Technology, University of Coimbra (FCTUC) in 2011, 2005 and 2000, respectively.
He is currently a Senior Lecturer at Nottingham Trent University (NTU) and a researcher at the Computational Neuroscience and Cognitive Robotics Research Group (CNCR). He is a member of the IEEE and the IEEE Robotics and Automation Society (RAS) since 2012 (member of the Technical Committee on Cognitive Robotics, T-CORO, since 2015, and member of the Technical Committee on Agricultural Robotics, TC AgRA, since 2019). Additionally, he is a member of the IEEE Life Sciences Community since 2013, the IEEE Systems, Man, and Cybernetics Society since 2015 and the IEEE Computational Intelligence Society since 2015. He is also a member of the British Computer Society since 2019. His main research interests are artificial perception and cognitive robotics.
\end{IEEEbiography} 
\vspace{-1cm}
\begin{IEEEbiography}
[{\includegraphics[width=1in,height=1.25in,clip,keepaspectratio]{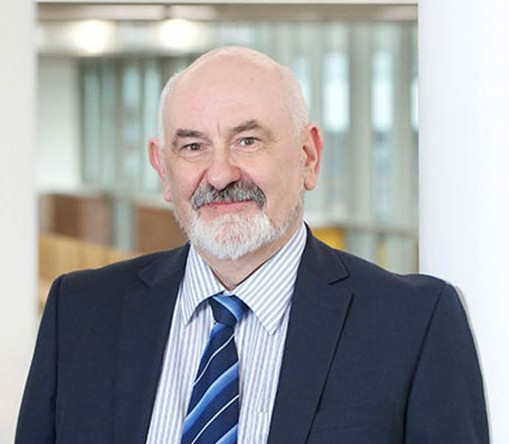}}]{Professor T. Martin McGinnity}
(SMIEEE, FIET) received a First Class (Hons.) degree in Physics in 1975, and a Ph.D degree in 1979. He currently holds a part-time Professorship in both Nottingham Trent University (NTU), UK and Ulster University (UU). He is the author or coauthor of 350+ research papers and leads the Computational Neuroscience and Cognitive Robotics research group at NTU. His current research is focused on the development of biologically-compatible computational models of human sensory systems, including auditory signal processing; human tactile emulation; human visual processing; sensory processing modalities in cognitive robotics; and implementation of neuromorphic systems on electronics hardware.
\end{IEEEbiography}
\EOD
\end{document}